\pdfoutput=1

\documentclass[11pt]{article}

\usepackage{emnlp2022}

\usepackage{times}
\usepackage{latexsym}

\usepackage[T1]{fontenc}

\usepackage[utf8]{inputenc}

\usepackage{microtype}

\title{Structurally Diverse Sampling for Sample-Efficient Training\\and Comprehensive Evaluation}

\author{\makecell{Shivanshu Gupta$^{1}$ ~~~~~~~ Sameer Singh$^{1,2}$  ~~~~~ Matt Gardner$^{3}$ } \\
$^{1}$University of California Irvine \hspace{4mm}
$^{2}$Allen Institute for AI\hspace{4mm}
$^{3}$Microsoft Semantic Machines  \hspace{4mm}   \\
\texttt{\makecell{\{shivag5,sameer\}@uci.edu, mattgardner@microsoft.com\\}}}

\usepackage{svg}
\usepackage{graphicx}
\usepackage{amssymb}
\usepackage{amsmath}
\usepackage{booktabs}
\usepackage{makecell}
\usepackage{multicol}
\usepackage{multirow}
\usepackage{enumitem,kantlipsum}
\usepackage{hyperref}
\usepackage{algorithm,algpseudocode}
\usepackage{arydshln}
\usepackage{caption}
\usepackage{bbm}
\usepackage{soul}
\usepackage{subcaption}

\newif\ifcomments
\commentstrue
\ifcomments
    \providecommand{\sg}[1]{{\protect\color{cyan}{[\textbf{shiv}: #1]}}}
    \providecommand{\sameer}[1]{{\protect\color{purple}{[\textbf{sameer}: #1]}}}
    \providecommand{\matt}[1]{{\protect\color{red}{[\textbf{matt}: #1]}}}
\else
    \providecommand{\sg}[1]{}
    \providecommand{\sameer}[1]{}
    \providecommand{\matt}[1]{}
\fi

\newcommand{\tightparagraph}[1]{\smallbreak\noindent\textbf{#1}}

\definecolor{applegreen}{rgb}{0.01, 0.65, 0.01}
\definecolor{cardinal}{rgb}{0.77, 0.12, 0.23}

\newcommand{\covr}{\textsc{COVR}}
\newcommand{\smcalflow}{\textsc{SM-CalFlow}}
\newcommand{\schemaqa}{\textsc{Schema2QA}}
\newcommand{\atis}{\textsc{Atis}}
\newcommand{\overnight}{\textsc{Overnight}}

\newcommand{\bigram}{\textsc{Bigram}}
\newcommand{\bigramfreq}{\textsc{Bigram[Freq]}}
\newcommand{\template}{\textsc{Template}}
\newcommand{\templatefreq}{\textsc{Template[Freq]}}
\newcommand{\strand}{\textsc{Subtree[RandEx]}}
\newcommand{\strandnewt}{\textsc{Subtree[RandNewT]}}
\newcommand{\stfreqnewt}{\textsc{Subtree[FreqNewT]}}

\begin{document}
\maketitle

\begin{abstract}

A growing body of research has demonstrated the inability of NLP models to generalize compositionally and has tried to alleviate it through specialized architectures, training schemes, and data augmentation, among other approaches. In this work, we study a different approach: training on instances with diverse structures.
We propose a model-agnostic algorithm for subsampling such sets of instances from a labeled instance pool with structured outputs.
Evaluating on both compositional template splits and traditional IID splits of 5 semantic parsing datasets of varying complexity, we show that \textit{structurally diverse training} using our algorithm leads to comparable or better generalization than prior algorithms in 9 out of 10 dataset-split type pairs.
In general, we find structural diversity to consistently improve sample efficiency compared to random train sets.
Moreover, we show that structurally diverse sampling yields comprehensive test sets that are a lot more challenging than IID test sets.
Finally, we provide two explanations for improved generalization from diverse train sets: 1) improved coverage of output substructures, and 2) a reduction in spurious correlations between these substructures.

\end{abstract}
\section{Introduction}

\begin{figure}[t]
    \centering
    \includegraphics[width=\linewidth]{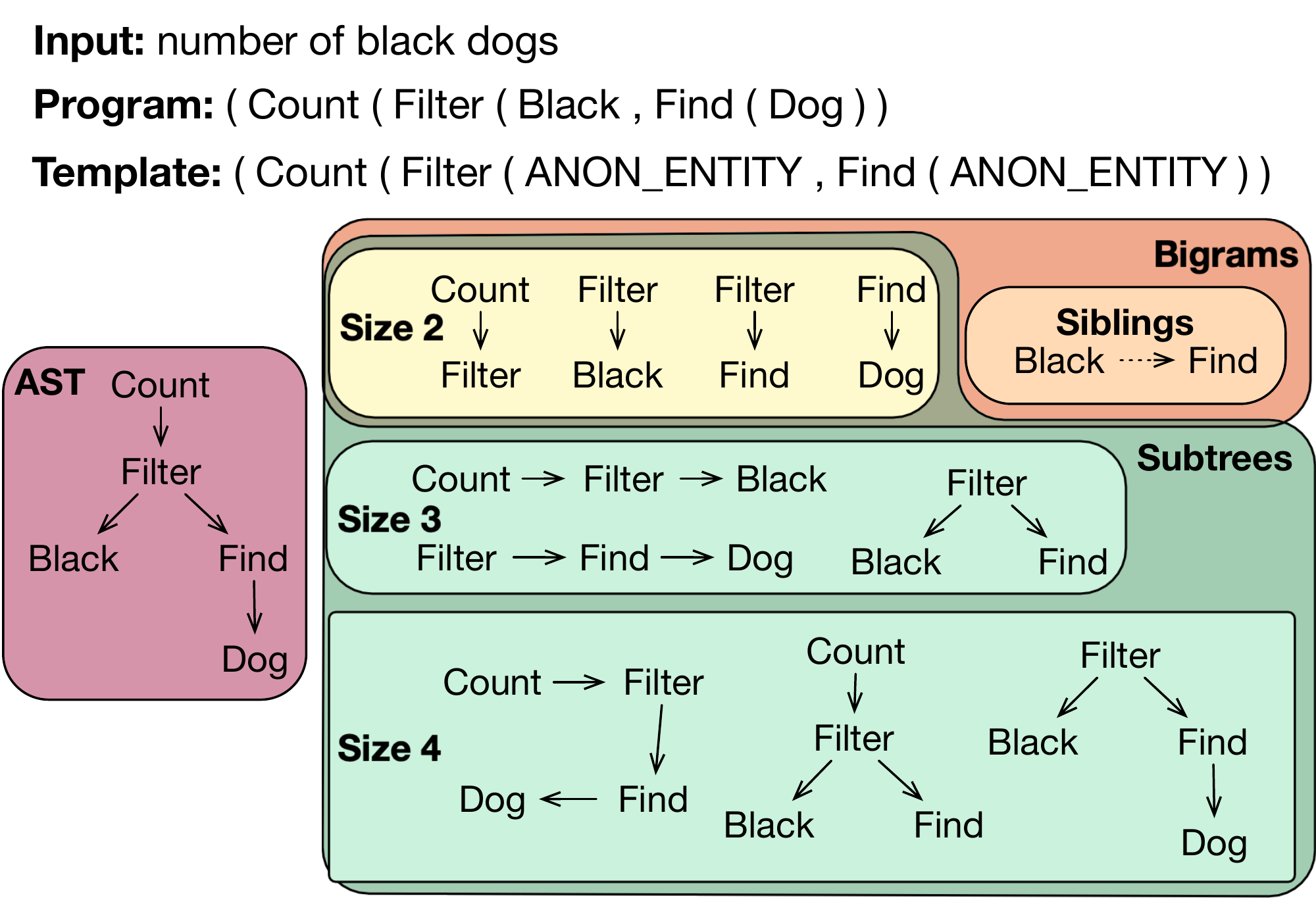}
    \caption{An instance from \covr{} and its abstract syntax tree (AST) and different types of substructures in it, including templates, bigrams, and subtrees. Note that subtrees will also include each node in the AST.}
    \label{fig:substructures}
\end{figure}

Systematic compositionality---expressing novel complex concepts as systematic compositions of expressions for simpler concepts---is the property underlying human languages' expressive power \cite{lake2017building,fodor1988connectionism}. However, NLP models struggle to generalize to novel composite expressions in the context of semantic parsing~\cite{lake2018scan,loula-etal-2018-rearranging,kim-linzen-2020-cogs}.

Data augmentation has been extensively explored for compositional generalization~\cite{Akyrek2021LearningTR,guo2021revisiting,wang-etal-2021-learning-synthesize,guo-etal-2020-sequence,qiu-etal-2021-csl}. However, instances in semantic parsing possess structure, such as abstract syntax trees (ASTs) of target programs (Figure \ref{fig:substructures}). Randomly selecting instances from a grammar, or even from human annotators, while ignoring similarity in their structure is likely to produce skewed distributions with spurious correlations between substructures. This \textit{task ambiguity} and \textit{underspecification} are known to cause overparameterized models to have high variance and poor generalization \cite{damour-etal-2020-underspecification,geirhos-etal-2020-shortcut-learning}. In contrast, a structurally diverse set of instances containing a variety of structures should better cover the combinatorial space of structures and yield more sample-efficient train sets. Moreover, this improved coverage should yield more comprehensive test sets than the traditional IID test sets.

Prior work on diverse sampling has shown improved sample efficiency in compositional (non-IID) splits, by
selecting instances to have diverse \emph{program templates}~\cite{oren-etal-2021-finding} or \emph{AST bigrams}~\cite{bogin2022unobserved} (see Figure~\ref{fig:substructures} for examples). However, for reasonably-sized programs, templates are very large structures, and bigrams very small; focusing on either does not necessarily improve diversity across a range of differently-sized structures. Moreover, there is little evidence of sample efficiency of diverse train sets in the IID setting and no exploration of diverse test sets.

In this study, we take a broader look at the advantages of structural diversity. We propose a general, substructure, and model-agnostic recipe for structurally diverse subsampling algorithms that interleaves selecting unseen substructures with selecting instances containing those substructures. Experimenting with different substructure and instance selection criteria, we show improved performance from 1) prioritizing the selection of more frequent unseen substructures in the training pool, 2) using subtrees\footnote{These could also be called subgraphs but we'll stick to ``subtree'' as the whole structures are trees.} as substructures that are neither too granular (bigrams) nor too large (templates), and 3) simultaneously diversifying over both small (subtree) and larger (template) substructures.

We evaluate on template and IID splits of five semantic parsing datasets of varying complexities and find that training on diverse train sets consistently outperforms random train sets. In particular, our proposed subtree diversity algorithm is the most consistent, performing comparably with or better than the template and bigram diversity in 9 out of 10 splits, with both bigram and template algorithms often performing worse than random subsampling in IID splits.
We study the efficacy of diversely sampled test sets and show that our diverse sampling algorithm yields test sets that are harder and more comprehensive than IID tests, especially with IID training sets. Finally, by 
comparing random and structurally diverse subsamples, we show that the latter 1) better cover the space of substructures, especially those in the long tail, and 2) have weaker spurious correlations between substructures, explaining their benefits in training and evaluation.

To conclude, our results demonstrate the effectiveness of structurally diverse train sets in inducing generalization and of structurally diverse test sets as comprehensive evaluations. We hope these insights will encourage the use of structural diversity as a criterion when sampling train or test sets from a pool of labeled instances. A setting where we expect our algorithm will be particularly useful is that of prototyping semantic parsers in zero-data settings, where a pool of programs is sampled from a grammar and mapped to canonical utterances \cite{wang-etal-2015-building,herzig-berant-2019-dont,campagna-etal-2020-zero,yin-etal-2022-ingredients}. Another potential application could be in exemplar selection for in-context learning. Our code, data, and models are available at \url{https://github.com/Shivanshu-Gupta/structural-diversity}.

\section{Setup}
\label{sec:motivation}
Semantic Parsing, the task we focus on, involves parsing an utterance $x$ into a program or logical form $y$. Following \citet{oren-etal-2021-finding} and \citet{bogin2022unobserved}, we will assume access to a pool of instances $D_{pool} = \{e_i, \ldots, e_n\}$, where $e_i = (x_i, y_i)$, from which we wish to sample instances. As the example in Figure \ref{fig:substructures} shows, logical forms possess hierarchical structure which can be represented as trees called \textit{abstract syntax trees} (ASTs). These structures are composed of smaller substructures such as subtrees of the AST, that are shared across logical forms. We can thus view each instance as a bag of output substructures: $\mathcal{C}(e_i) = \mathcal{C}(x_i, y_i)  = \mathcal{C}(y_i) =  \{c^{\{i\}}_1, \ldots, c^{\{i\}}_m\} \subseteq C$, where $\mathcal{C}$ is the mapping from instances to their substructures and $C$ is the set of all unique substructures.

The goal of structurally-diverse sampling is to select instances that contain among them a variety of different substructures.
Our hypotheses, that we verify in \S\ref{sec:analysis}, are that a collection of such instances would both better cover the combinatorial space of structures and reduce spurious correlations between substructures compared to random instances.
A structurally diverse train set should thus give the model more information about the \textit{system} underlying the instances' structure and hence improve generalization.
Similarly, its improved coverage should also make a structurally diverse test set more comprehensive.
This is analogous to how one would choose to evaluate on a test set with balanced classes even when the training data might have a considerable imbalance.

\section{Structurally Diverse Subsampling}
\label{sec:algorithm}

\subsection{Substructures}
\label{sec:substruct}
Prior work~\citep{oren-etal-2021-finding,bogin2022unobserved} has used program templates and bigrams in program ASTs (Figure \ref{fig:substructures}) for their diverse subsampling algorithms. Bigrams are pairs of parent-child and sibling nodes in program ASTs. Program templates represent the reasoning pattern in the program and are obtained by replacing certain program tokens such as strings and numbers with their abstract type. Their exact implementation depends on the dataset and is described in \S\ref{sec:datasets}.

While we follow \citet{oren-etal-2021-finding} and \citet{bogin2022unobserved} in extracting substructures from programs alone, we choose to use different substructures, as templates and bigrams do not seem to hve the optimal granularity to diversify over.
Templates are too coarse; they still share a lot of structure, and there may be a combinatorially large number of them. On the other hand, bigrams are too fine and may be unable to capture many salient structural patterns.
We thus use subtrees of the program AST up to a size $d$. The ASTs are constructed as in \citet{bogin2022unobserved}: tokens in a program are categorized as either functions, values, or structural tokens (such as parentheses or commas) that define the hierarchical structure of the program.

\subsection{Algorithm}\label{sec:recipe}
\label{sec:algo}

As motivated in \S\ref{sec:motivation}, we want to sample instances containing among them a variety of substructures. Additionally, we wish to experiment with: 1) prioritizing substructures more common in the pool, as we expect it to reduce the divergence of the subsample from the pool distribution; and 2) simultaneously diversifying over both fine (subtree) and coarse (template) granularity substructures.
One approach would be constructing an optimization problem that directly selects an optimally diverse set of instances. However, this has the challenge of defining a measure of diversity that is also tractable to optimize for large pools and a large number of substructures (see Table \ref{tab:ds-stats}). Additonally, it would be harder to experiment with the different variations of diverse sampling described above.

We thus take the route of an iterative algorithm (pseudo-code in Algorithm \ref{alg:algo}) that alternates between picking a substructure and picking an instance with that substructure till the requisite number, $B$, of instances has been sampled. Here, $w_c$ and $w_e$ specify the substructure and instance selection criteria by assigning weights based on the current state comprising sampled instances $D_{sample}$, substructures $C_{sample}$, and templates $T_{sample}$. The algorithm resets $C_{sample}$ and $T_{sample}$ once all substructures and templates, respectively have been sampled to allow cycling over them repeatedly. In the following sections, we will show that with different substructure definitions and substructure and instance-weighting schemes, Algorithm \ref{alg:algo} can subsume both the template diversity and bigram diversity algorithms from \citet{oren-etal-2021-finding} and \citet{bogin2022unobserved}. Additionally, it will allow us to experiment with the variations described above.

\begin{algorithm}[tb]
    \caption{Structurally Diverse Subsampling}\label{alg:algo}
    \small
    \begin{algorithmic}
        \Require Instance pool $D_{pool}$; set of all substructures C, and templates T in pool: instance-to-substructure mapping $\mathcal{C}$; template mapping $\mathcal{T}$; substructure weight function $w_c$; instance weight function $w_e$; training budget $B$
        \State $D_{sample},C_{sample},T_{sample} \gets \phi$
        \State $i \gets 0$
        \While{$i < B$}
            \State $c = \operatornamewithlimits{argmax}\limits_{c \in C} w_c(c, D_{pool}, D_{sample}, C_{sample})$
            \State $e = \operatornamewithlimits{argmax}\limits_{\substack{e \in D_{pool} \\ \text{s.t. } c \in \mathcal{C}(e)}} w_e(e, D_{pool}, D_{sample}, T_{sample})$
            \State $D_{pool} \gets D_{pool} \setminus {e}$
            \State $D_{sample} \gets D_{sample} \cup {e}$
            \State $C_{sample} \gets C_{sample} \cup {c}$
            \State $T_{sample} \gets T_{sample} \cup \{\mathcal{T}(e)\}$
            \State $C \gets \bigcup\limits_{e \in D_{pool}} \mathcal{C}(e)$
            \State $T \gets \{\mathcal{T}(e): e \in D_{pool}\}$

            \If{$C_{sample} = C$}
                \State $C_{sample} \gets \phi$
            \EndIf
            \If{$T_{sample} = T$}
                \State $T_{sample} \gets \phi$
            \EndIf
        \EndWhile
        \State \textbf{return} $D_{sample}$
    \end{algorithmic}
\end{algorithm}

\subsection{Subtree Diversity}
\label{sec:subtree}

Our proposed subsampling algorithm diversifies over subtrees as defined in \S\ref{sec:substruct}, i.e., $\mathcal{C}(x, y)$ is the set of subtrees of size $\leq d$ in the AST of $y$. We will use $d = 4$. Since we want to prioritize selecting more frequent unsampled substructures in the pool, we will use $w_c(c) = \mathbbm{1}[c \notin C_{sample}]F_{pool}(c)$ where $F_{pool}(c)$ is the number of instances in the pool containing $c$.

For selecting instances given a sampled substructure, we use these instance weighting schemes:

\begin{enumerate}[nosep]
    \item \textsc{RandEx} samples an instance uniformly at random: $w_e(e)$ is a constant.
    \item \textsc{RandNewT} samples an instance with unseen template: $w_e(e) = \mathbbm{1}[\mathcal{T}(e) \notin T_{sample}]$.
    \item \textsc{FreqNewT} samples instance with the most frequent unsampled template: $w_e(e) = \mathbbm{1}[\mathcal{T}(e) \notin T_{sample}]F_{pool}(\mathcal{T}(e))$ where $\mathcal{T}(e)$ is $e$'s template and $F_{pool}(t)$ is the number of instances in the pool with template $t$.
\end{enumerate}
The last two schemes seek to improve the coverage of templates to allow the diversifying of both subtrees and templates.

\subsection{Template and Bigram Diversity}

\tightparagraph{Template Diversity} from \citet{oren-etal-2021-finding}, henceforth referred to as \template{}, can be implemented in the framework of Algorithm \ref{alg:algo} as:
\begin{itemize}[nosep]
    \item $\mathcal{C}(x, y)$ is a singleton set containing the template for $y$.
    \item $w_c(c)$ is a constant function i.e. a random template.
    \item $w_e(e)$ is a constant function (\textsc{RandEx}).
\end{itemize}
Additionally, we also experimented with selecting only among the unsampled templates i.e. $w_c(c) = \mathbbm{1}\left[c \notin C_{sample}\right]$ and with prioritizing more frequent unsampled templates using the substructure weighting scheme of \S\ref{sec:subtree}. We found both of these to improve performance but only include the results for the latter. We will refer to it as \templatefreq{}.

\tightparagraph{Bigram Diversity} from \citet{bogin2022unobserved}
, hereafter referred to as \bigram{}, can also be implemented in the framework of Algorithm \ref{alg:algo} as:

\begin{itemize}[nosep]
    \item $\mathcal{C}(x, y)$ is the set of bigrams in $y$'s AST as defined by \citet{bogin2022unobserved}.
    \item \citet{bogin2022unobserved}'s bigram diversity algorithm randomly samples from unsampled bigrams until there is still an unsampled bigram and then any random bigram. This can be formulated as: $w_c(c) = \mathbbm{1}\left[c \notin \tilde{C}_{sample}\right]$ where $\tilde{C}_{sample} = \bigcup\limits_{e \in D_{sample}} \mathcal{C}(e)$ is the set of all bigrams in the current sample.
    \item $w_e(e)$ is a constant function (\textsc{RandEx}).
\end{itemize}
We also experimented with bigram diversity with the substructure weighting scheme of \S\ref{sec:subtree}. We will refer to this algorithm as \bigramfreq{}.

\section{Experiments}
\label{sec:exp}
Given a dataset $\mathcal{D}$ of utterance-program pairs, we create three different types of splits with each split consisting of a training pool $D_{pool}$ and test set $D_{test}$. We then compare the various subsampling algorithms described in \S\ref{sec:algorithm} by using them to sample training sets $D_{train}$ of varying budget size $B$ from $D_{pool}$ and evaluating on $D_{test}$.

\subsection{Splits}
\label{sec:splits}

\tightparagraph{IID split} For this we randomly sample instances from $\mathcal{D}$ to use as $D_{test}$, keeping the rest for $D_{pool}$.

\tightparagraph{Template split} This is a type of compositional split proposed by \cite{finegan-dollak-etal-2018-improving}. Here instances are grouped based on their program template as described in \S\ref{sec:substruct}
The split is then created by randomly splitting the set of templates into a train set and a test set and using examples for train or test templates as $D_{pool}$ or $D_{test}$ respectively. We follow the procedure of \citet{bogin2022unobserved} to obtain solvable template splits where every token in the test set also occurs in the train set.

\tightparagraph{Subtree split} In \S\ref{sec:motivation} we argued that diversely subsampled sets of instances should also make for more comprehensive test sets. We thus experiment with a third type of split: we use \stfreqnewt{} diverse subsampling to sample test sets $D_{test}$ from $\mathcal{D}$, keeping the rest as $D_{pool}$.

\subsection{Datasets}
\label{sec:datasets}

\begin{table*}
\centering
\small
\begin{tabular}{p{1.75cm}p{4.4cm}p{8.5cm}}
\toprule
{\bf Dataset} & {\bf Input Utterance} & {\bf Target Program}\\
\midrule
\multirow{2}{*}{\makecell[l]{\covr{}\\(synthetic)}} & \textit{What is the number of black dog that is chasing mouse ?} & \multirow{2}{*}{\makecell[l]{\texttt{count ( with\_relation ( filter ( black , find ( dog ) ) , } \\ \texttt{chasing , find ( mouse ) ) )}}}\\

\addlinespace
\multirow{2}{*}{\makecell[l]{\overnight{}\\(synthetic\textsuperscript{$\dagger$},\\
natural\textsuperscript{$\circ$})
}} & \textsuperscript{$\dagger$}\textit{person whose height is 180 cm and whose birthdate is 2004} & \multirow{2}{*}{\makecell[l]{\texttt{(listValue (filter (filter (getProperty (singleton en.person)} \\ \texttt{(string !type)) (string height) (string =) (number 180 en.cm))} \\ \texttt{(string birthdate) (string =) (date 2004 -1 -1)))}}}\\
 & $\circ$\textit{what person born in 2004 is 180 cm tall} & \\

\addlinespace
\makecell[l]{\schemaqa{}\\(synthetic)} & \makecell[l]{\textit{what people are named aideliz li}} & 
\makecell[l]{\texttt{( Person ) filter id =~ "aideliz li"}}\\

\addlinespace
\multirow{2}{*}{\makecell[l]{\atis{}\\(natural)}} & \textit{a flight on continental airlines leaving boston and going to denver} & \multirow{2}{*}{\makecell[l]{\texttt{( lambda \$0 e ( and ( flight \$0 ) ( airline \$0 co : al ) } \\ \texttt{( from \$0 boston : ci ) ( to \$0 denver : ci ) ) )}}}\\

\addlinespace
\multirow{2}{*}{\makecell[l]{\smcalflow{}\\(natural)}} & \textit{When is my next staff meeting scheduled for?} & \multirow{2}{*}{\makecell[l]{\texttt{(Yield (Event.start (FindNumNextEvent (Event.subject?} \\ \texttt{(?~= "staff meeting")) 1L)))}}}\\
\bottomrule
\end{tabular}
\caption{Examples of input utterance and target program pairs for the datasets used in this work.}
\label{tab:ds-samples}
\end{table*}

\begingroup
\setlength{\tabcolsep}{3pt} %
\begin{table}
\centering
\small
\begin{tabular}{lrrrr}
\toprule
  Dataset &  Instances &  Bigrams &  Subtrees &  Templates \\
\midrule
     \atis &       5037 &     5091 &     41536 &       1149 \\
     \covr &     100000 &      298 &      4490 &      29141 \\
\overnight &       4419 &      354 &      3015 &         87 \\
\smcalflow &     106072 &    43689 &    208527 &      21082 \\
\schemaqa &    1577860 &      223 &      3060 &        139 \\
\bottomrule
\end{tabular}
\caption{Number of instances, bigrams, subtrees (size $\leq 4$), and templates in the datasets used for structural diversity experiments.}
\label{tab:ds-stats}
\end{table}
\endgroup

We use five semantic parsing datasets from diverse domains and complexity for our analysis, with both synthetic and natural language input utterances. Tables \ref{tab:ds-samples} and \ref{tab:ds-stats} show a few examples and statistics regarding number of instances and different types of substructures.

\noindent\textbf{COVR}: A synthetic dataset that uses a variable-free functional query language and is generated using a synchronous context-free grammar (SCFG) adapted from the VQA dataset of \citet{bogin-etal-2021-covr}. We use the SCFG to generate 100K examples for our experiments.

\noindent\textbf{ATIS} \cite{Hemphill1990atis,dahl-etal-1994-expanding}: A dataset of natural language queries about aviation paired with $\lambda$-calculus programs.

\noindent\textbf{Overnight} \cite{wang-etal-2015-building}: A dataset containing both synthetic and natural language utterances from 11 domains (e.g. \textit{socialnetwork, restaurants}, etc.) paired with Lambda-DCS logical forms.

\noindent\textbf{Schema2QA} \cite{xu2020schema2qa}: Uses the ThingTalk language \cite{campagna2019genie}. We use the synthetic instances from the \textit{people} domain generated by \citet{oren-etal-2021-finding}.

\noindent\textbf{SM-CalFlow} \cite{andreas-etal-2020-task}: Consists of dialogs paired with LISP programs. Each instance is a single dialogue turn from one of two domains about creating calendar events or querying an org chart.

Except for SM-CalFlow, which we took from \citet{andreas-etal-2020-task}, we used the preprocessed versions of the above datasets provided by \citet{bogin2022unobserved} and used their code\footnote{\url{https://github.com/benbogin/unobserved-local-structures}} to anonymize programs and produce ASTs. For SM-CalFlow, we anonymized strings and numbers such as ``staff meeting'' and ``1L'' in Table \ref{tab:ds-samples} and used nltk\footnote{\url{https://www.nltk.org/}} to produce ASTs. Additionally, we excluded all instances with a particular program that accounted for more than 12\% of the original dataset.

\subsection{Model and Training}
\label{sec:model}

We use the pre-trained BART-base model \cite{lewis-etal-2020-bart} for our experiments, fine-tuning it on subsamples for each dataset, split, and subsampling algorithm. For each dataset, we use 4 different random seeds to create 4 splits of each type. Then for each split and training budget, we use 3 different seeds to subsample 3 training sets for each subsampling algorithm. See App. \ref{app:training} for more details.

\label{sec:results}
\begin{figure*}[tb]
    \centering
    \includegraphics[width=\textwidth]{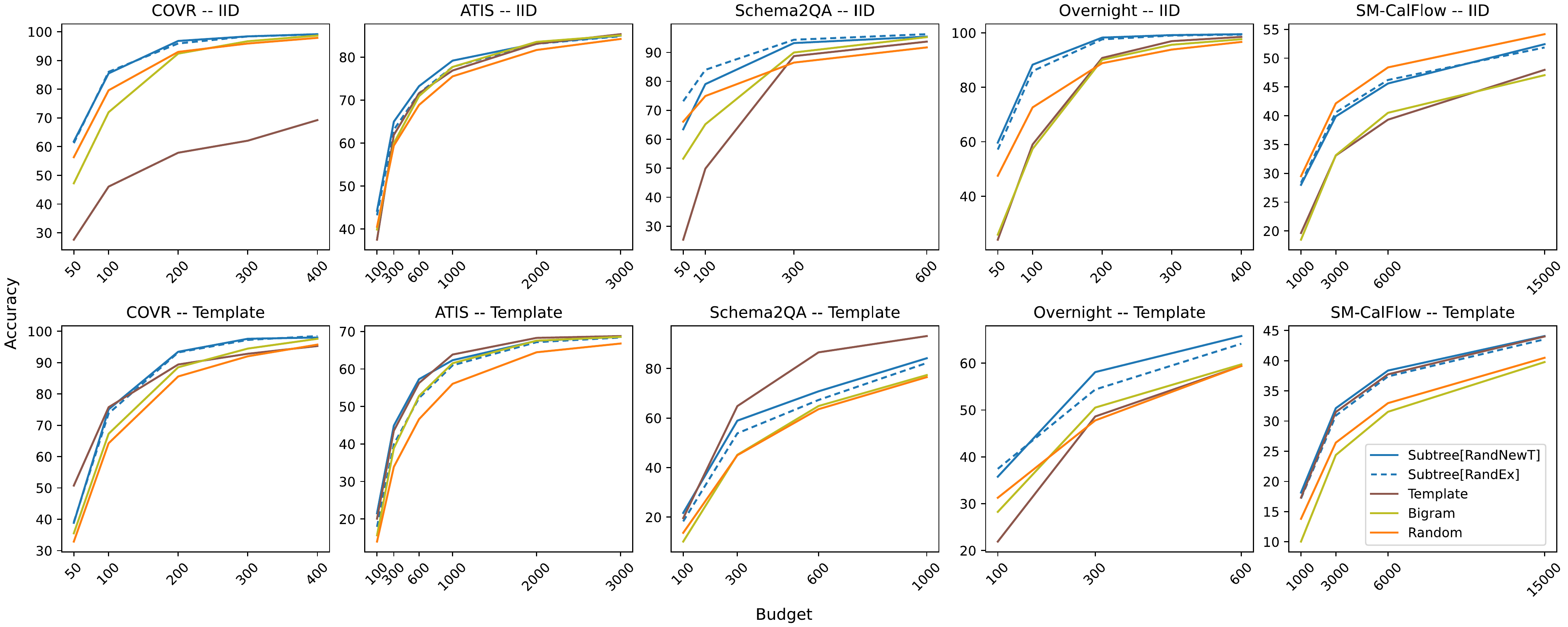}
    \caption{
    Comparing different subsampling algorithms on IID and Template splits for varying budgets.
    }
    \label{fig:efficiency}
\end{figure*}

\begin{figure*}[tb]
    \centering
    \includegraphics[width=\textwidth]{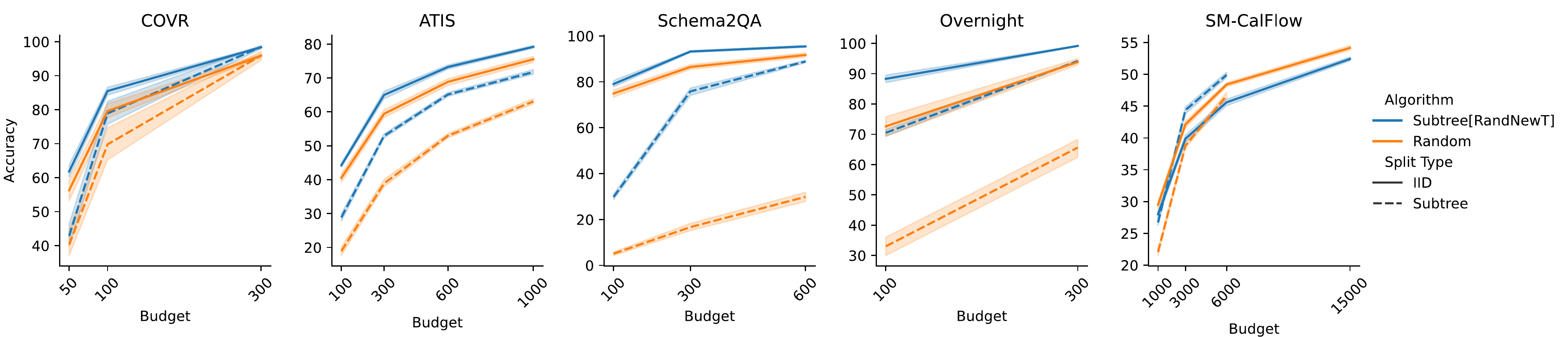}
    \caption{
        Subtree splits are more comprehensive than IID splits and hence are much harder with randomly sampled train sets but not with diverse train sets.
        }
    \label{fig:iid_subtree_all_trn}
\end{figure*}

\section{Results}

\subsection{Structurally Diverse Train Sets}

Figure \ref{fig:efficiency} compares the various structurally diverse sampling algorithms, and random sampling on Template and the IID splits of the different datasets.

\tightparagraph{Structural Diversity improves sample efficiency.} Random sampling is outperformed by structurally diverse sampling in 9 out of 10 dataset-split type combinations with about 2x sample efficiency in template splits of all five datasets except \covr{} and IID splits of \schemaqa{} and \overnight{}. The only exception is the IID split of \smcalflow{}. We believe this is due to a greater imbalance in substructures in \smcalflow{} where over a third of templates (as well as bigrams and subtrees) only appear in a single instance.

\tightparagraph{Subtree diversity is most consistent.} 
Our proposed subtree diversity algorithm is the most consistent among the various diverse sampling algorithms. In all but 1 of the 10 splits, it outperforms or matches both template and bigram diversity which often lag behind even random sampling. The only exception is template diversity in the template split of \schemaqa{}, which has very few templates (see Table \ref{tab:ds-stats}).
These results suggest that the optimal substructure granularity depends on the task. For datasets with little structural diversity, diversifying over templates is sufficient. However, diversifying over smaller substructures is more beneficial for complex datasets with many program templates overlapping in structure as it exploits the structural similarity of programs.

\tightparagraph{Combining subtree and template diversity improves performance.} Comparing \strandnewt{} and \strand{} we see that the former performs better in template splits with minor reductions in IID splits. This suggests that simultaneously diversifying over fine and coarse substructures, subtrees, and templates here, can be more effective than either alone.

\tightparagraph{Prioritizing more frequent compounds improves efficiency.} Figure \ref{fig:prior-efficiency} compares \bigram{} and \template{} with \bigramfreq{} and \templatefreq{} respectively. It is evident that on IID splits, bigram and template diversity greatly benefit from prioritizing more frequent substructures instead of random selection, with only minor degradation in template splits.

\subsection{Structurally Diverse Test Sets}

Figure \ref{fig:iid_subtree_all_trn} compares the performance of random and diverse (\strand{}) subsamples in IID and Subtree splits. 
It is evident that Subtree splits are more challenging than IID splits. 
More importantly, this split is much harder with random train sets than with diverse train sets, which, as the training budget increases, quickly close the gap with IID performance on all datasets except \atis{}. 
This is expected since, unlike template splits, this split is challenging not because of a systematic distributional gap between train and test sets, but because it tests for a lot more structures that a random train set may not cover. 
We thus believe that structurally diverse test sets also enable more comprehensive evaluation.

\section{Analysis}
\label{sec:analysis}

Having seen that structurally diverse datasets improve generalization, we now investigate why they do so. 
In \S\ref{sec:motivation} we hypothesize that structural diversity has two benefits over random sampling: (1) improved coverage of the space of substructures, and (2) reduced spurious correlations. We now evaluate whether this is indeed the case.

We will view each instance as the set of subtrees contained in it, i.e., $e_i \subset C$ where $C$ is the set of all subtrees in the dataset. We define the frequency of a subtree, $s$, in a set of instances $A$ as the number of instances containing it, i.e. $F_{A}(s) = |\{s \in \mathbf{e}: \mathbf{e} \in A\}|$.
The frequency of a pair of subtrees, $s_i, s_j$, is analogously defined as $F_{A}(s_i, s_j) = |\{s_i, s_j \in \mathbf{e}: \mathbf{e} \in A\}|$. The rank of a subtree in $A$ is defined as its position in a list of all subtrees descending in frequency, with equally-frequent subtrees assigned the average of their ranks.

\subsection{Improved Coverage of the Long tail}\label{sec:coverage}

\begin{figure*}[tb]
    \centering
    \includegraphics[width=\textwidth]{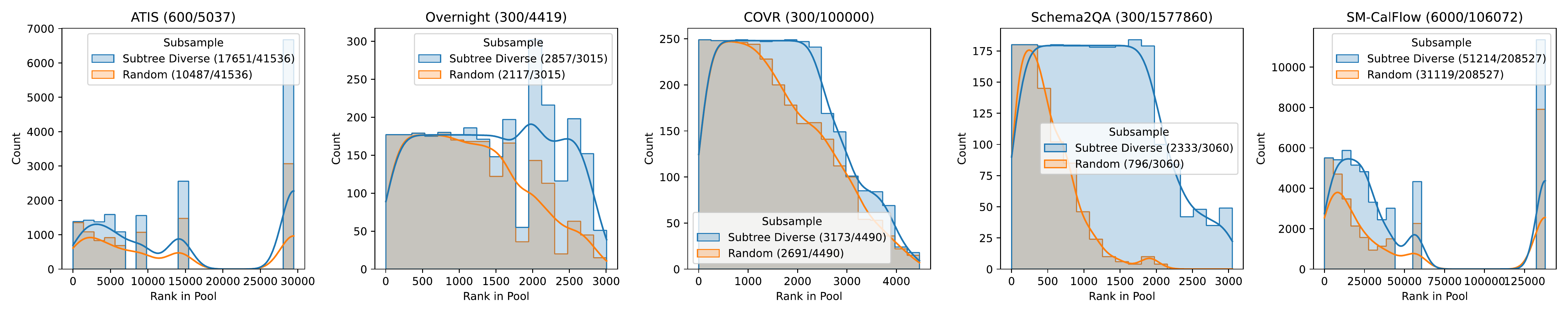}
    \caption{
        The number of unique subtrees in diverse and random subsamples bucketed by the rank of their frequency in the pool. Titles include the size of the subsamples and the pool, and the legends include the number of unique subtrees in the subsamples v/s the pool.
    }
    \label{fig:subtree-coverage}
\end{figure*}

\begin{figure*}[tb]
    \centering
    \includegraphics[width=\textwidth]{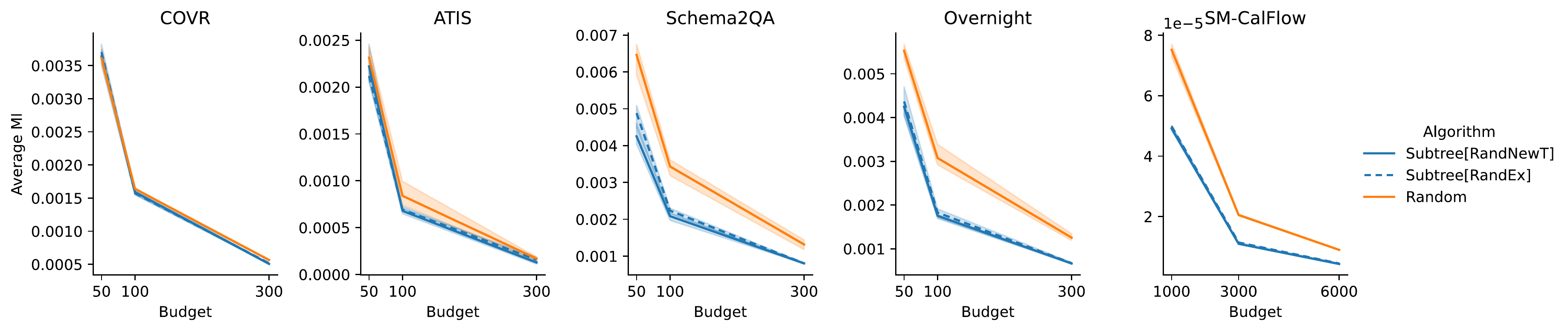}
    \caption{Average Mutual Information (AMI) of random subsamples compared with that of diverse subsamples.}
    \label{fig:diverse-corr}
\end{figure*}

To verify whether structural diversity improves the coverage of substructures, we use random and \strand{} subsampling to take equal-sized subsamples from pools consisting of entire datasets, $\mathcal{D}$, and compare the distribution of subtrees.
Figure \ref{fig:subtree-coverage} shows the number of unique subtrees in random and diverse subsamples bucketed by rank in $\mathcal{D}$. As can be seen in the figure, structurally diverse sampling improves the coverage of substructures, especially those in the long tail. Crucially, because substructures are chosen by frequency, this coverage of the long tail does not come at the expense of frequent substructures.
We refer the reader to Appendix \ref{app:coverage} for additional analyses showing that although \strand{} sampling directly diversifies over only output subtrees, it also improves coverage of program templates as well as n-grams in input utterances.

\subsection{Weakened Spurious Correlations}\label{sec:correlation}

Structurally diverse datasets might improve generalization by weakening the spurious correlations between substructures that would exist in any dataset of bounded size.

\tightparagraph{Measure for Correlations}
We measure correlations between substructures in a set of instances $D$ as the sum of pairwise mutual information (MI) between them, normalized to account for the larger number of distinct substructures in diverse subsamples. Let $S = {s_1, \ldots, s_m}$ be the set of all subtrees in a sample. We define an indicator random variable $I_s$ for each subtree $s$ that is 1 if $s$ is present in a random instance, 0 otherwise. MI between two subtrees, $s_i$ and $s_j$ is then defined as,
\begin{equation}
    \operatorname{MI}[s_i, s_j] = -\sum\limits_{I_{s_i}, I_{s_j}} \tilde{p}(I_{s_i}, I_{s_j}) \log \frac{\tilde{p}(I_{s_i}, I_{s_j})}{\tilde{p}(I_{s_i})\tilde{p}(I_{s_j})}
\end{equation}
where $\tilde{p}(I_s=1)$ and $\tilde{p}(I_{s_i}=1, I_{s_j}=1)$ are empirical probabilities defined as,
\begin{align}
    \tilde{p}(s=1) &= \frac{|\{s \in \mathbf{e}: \mathbf{e} \in D\}|}{|D|} \\
    \tilde{p}(s_i=1, s_j=1) &= \frac{|\{s_i, s_j \in \mathbf{e}: \mathbf{e} \in D\}|}{|D|}
\end{align}
Other probabilities are defined analogously. Finally, our measure of correlations is the average mutual information (AMI) across all subtree pairs:
\begin{equation}
    \operatorname{AMI}(D) = \frac{1}{|S|^2}\sum\limits_{i, j} \operatorname{MI}[s_i, s_j]
\end{equation}

\begingroup
\setlength{\tabcolsep}{3pt} %
\begin{table}
\centering
\small
\begin{tabular}{lcccc}
\toprule
Split Type & Budget & COVR &  Overnight &  Schema2QA \\
\midrule
IID & 100 & -0.50 &      -0.83 &      -0.39 \\
         & 300 & -0.70 &      -0.76 &      -0.56 \\
Template & 100 & -0.57 &      -0.58 &      -0.68 \\
         & 300 & -0.60 &      -0.56 &      -0.64 \\
\bottomrule
\end{tabular}
\caption{Spearman correlations of AMI and Accuracy for train sets sampled randomly and using \strand{} and \strandnewt{} algorithms averaged across 4 split seeds (from \S~\ref{sec:exp}).}
\label{tab:corr}
\end{table}
\endgroup

\tightparagraph{Results} For each dataset, we took 3 samples using random and subtree subsampling algorithms, treating the entire dataset as pool, and computed the AMI over their substructures.
Figure \ref{fig:diverse-corr} shows that diverse subsamples have lower AMI than random subsamples across different training budgets and the different split types, confirming our hypothesis that structurally diverse datasets have lower correlation.
Moreover, examining Spearman correlations of AMI and accuracy (Table \ref{tab:corr}) for train sets subsampled using random, \strand{} and \strandnewt{} subsampling algorithms from the experiments in \S\ref{sec:exp}, we find that they are significantly negatively correlated, giving further credence to our measure.
These results substantiate our hypothesis that reducing spurious correlations is indeed one way diverse subsamples improve generalization.

\section{Related Work}
\label{sec:related}

\tightparagraph{Generalization} Work on improving compositional generalization has considered many approaches including specialized architectures \cite{herzig-berant-2021-span,bogin-etal-2021-latent,chen2020compositional,Gordon2020Permutation,yin-etal-2021-compositional}, data augmentation \cite{andreas-2020-good,Akyrek2021LearningTR,guo2021revisiting}, modifications to training methodology \cite{oren-etal-2020-improving,csordas-etal-2021-devil}, and meta learning \cite{conklin-etal-2021-meta,lake-2019-metaseq2seq}. Of these, data-augmentation has the advantage of being model-agnostic; however, as argued in this work, randomly selecting training instances is inefficient. Our work builds upon \citet{oren-etal-2021-finding} and \citet{bogin2022unobserved} in arguing for structurally diverse training.

\tightparagraph{Training} Many methods have been explored for training instance selection, including: 1. Active Learning (AL) \cite{LEWIS1994148,settles-craven-2008-analysis}, where instances are iteratively selected for annotation based on criteria such as model uncertainty \cite{yarin-bayesian-active-learning} or diversity \cite{sener-core-set}; and 2. adversarial selection of instances that model fails on \cite{bras-etal-2020-adversarial,sakaguchi-etal-2021-winogrande,wallace-etal-2019-trick,nie-etal-2020-adversarial,kiela-etal-2021-dynabench}. However, given that these methods use a model in the loop for selection and generally work in the input domain, they are not comparable to our method. Additionally, the coupling of the dataset and model, and the lack of generalizability of AL heuristics has been shown to limit the effectiveness of active learning in practice \cite{lowell-etal-2019-practical,karamcheti-etal-2021-mind}.
\citet{tamkin-2022-active-learning} recently showed that pretrained models benefit more from AL by preferring ambiguous instances or ones with uncommon feature(s) thereby reducing task under-specification which can lead to higher variance and instability in models under-constrained by their training datasets \cite{damour-etal-2020-underspecification,geirhos-etal-2020-shortcut-learning}. Our analyses from \S\ref{sec:analysis} show that structurally diverse sampling also attempts to reduce under-specification albeit by directly sampling from a pool model-agnostically.

\tightparagraph{Evaluation}
Out-of-distribution (OOD) tests are essectial as testing in-distribution may not penalize models for learning spurious patterns \cite{linzen-2020-accelerate}. Our work on sampling diverse test sets is most related to methods for creating splits to test for compositional generalization from existing datasets \cite{keysers2020measuring, shaw-etal-2021-compositional,bogin2022unobserved}. However, unlike these, diverse test sets do not attempt to create a systematic gap between the train and test sets, but, as shown in \S\ref{sec:coverage}, cover the space of structures better.

\section{Conclusion}
\label{conclusion}

In this work, we studied the benefits of structural diversity for the task of semantic parsing. We proposed a novel model-agnostic algorithm for sampling structurally diverse instances by diversifying over subtrees in program ASTs. Evaluating on multiple datasets with varying complexities and on both IID and compositional splits, we demonstrated that diversity almost always, and often significantly, improves generalization. We further demonstrated that structural diversity also yields more comprehensive test sets than traditional IID test sets. Finally, we showed that these benefits of structural diversity are likely a manifestation of improved coverage of the long-tail of substructures as well as a reduction in spurious correlations between them.

We hope that our results demonstrating the importance of structural diversity will encourage future research on better structurally diverse sampling algorithms as well as their use to guide train and test set construction. The algorithms in this work are applicable whenever a large pool of possible output programs is available. One such setting is of prototyping a semantic parser in a zero-data setting where a pool of outputs can be sampled from a grammar and mapped to canonical utterances that can subsequently be paraphrased for linguistic variation either manually or automatically \cite{wang-etal-2015-building,campagna-etal-2020-zero}. Another scenario is that of selecting instances for in-context or few-shot learning \cite{brown-etal-2020-fewshot}.

Perhaps more importantly, our results indicate that the failure of NLP models on compositional generalization may be due in part to the lack of diversity in them, suggesting the need to create more diverse benchmarks.

\section*{Acknowledgements}

We would like to thank the anonymous reviewers for their feedback. Further we'd also thank Ben Bogin, Yasaman Razeghi, and Catarina Belem for their useful comments. This work was sponsored in part by the DARPA MCS program under Contract No. N660011924033 with the United States Office Of Naval Research, in part by funding by AI2, and in part by the NSF grant \#IIS-1817183. The views expressed are those of the authors and do not reflect the policy of the funding agencies.

\section*{Limitations}

A key limitation of the algorithms discussed in this work is their limited applicability due to the assumption of the availability of a pool of output programs paired with input utterances (natural or canonical). To alleviate these, future work can look at diverse sampling directly from a grammar without the intermediate step of sampling a pool or by preferring structural diversity in the input space instead of the output space i.e. on natural language utterances. The latter will additionally allow these methods to be applied to linguistic tasks other than semantic parsing.

Additionally, we took the route of a greedy iterative sampling algorithm to allow us to tractably subsample from large pools with large number of substructures. Future work can thus explore tractable optimization of some measure of diversity of sets of instances.

\section*{Ethics Statement}
In this work, we studied the efficacy of structurally diverse sampling when creating training and test sets for semantic parsing. Since we focus on structures derived from output programs and not language utterances it is unlikely to have any direct social impact or induce any biases in NLP systems.

\bibliography{bibliography/anthology-2022-rebiber,bibliography/anthology-2021-rebiber,bibliography/anthology-2020-rebiber,bibliography/anthology-2019-rebiber,bibliography/anthology-2018-rebiber,bibliography/custom-rebiber}

\begin{thebibliography}{52}
\expandafter\ifx\csname natexlab\endcsname\relax\def\natexlab#1{#1}\fi

\bibitem[{Aky{\"{u}}rek et~al.(2021)Aky{\"{u}}rek, Aky{\"{u}}rek, and
  Andreas}]{Akyrek2021LearningTR}
Ekin Aky{\"{u}}rek, Afra~Feyza Aky{\"{u}}rek, and Jacob Andreas. 2021.
\newblock \href {https://openreview.net/forum?id=PS3IMnScugk} {Learning to
  recombine and resample data for compositional generalization}.
\newblock In \emph{9th International Conference on Learning Representations,
  {ICLR} 2021, Virtual Event, Austria, May 3-7, 2021}. OpenReview.net.

\bibitem[{Andreas(2020)}]{andreas-2020-good}
Jacob Andreas. 2020.
\newblock \href {https://doi.org/10.18653/v1/2020.acl-main.676} {Good-enough
  compositional data augmentation}.
\newblock In \emph{Proceedings of the 58th Annual Meeting of the Association
  for Computational Linguistics}, pages 7556--7566, Online. Association for
  Computational Linguistics.

\bibitem[{Andreas et~al.(2020)Andreas, Bufe, Burkett, Chen, Clausman, Crawford,
  Crim, DeLoach, Dorner, Eisner, Fang, Guo, Hall, Hayes, Hill, Ho, Iwaszuk,
  Jha, Klein, Krishnamurthy, Lanman, Liang, Lin, Lintsbakh, McGovern,
  Nisnevich, Pauls, Petters, Read, Roth, Roy, Rusak, Short, Slomin, Snyder,
  Striplin, Su, Tellman, Thomson, Vorobev, Witoszko, Wolfe, Wray, Zhang, and
  Zotov}]{andreas-etal-2020-task}
Jacob Andreas, John Bufe, David Burkett, Charles Chen, Josh Clausman, Jean
  Crawford, Kate Crim, Jordan DeLoach, Leah Dorner, Jason Eisner, Hao Fang,
  Alan Guo, David Hall, Kristin Hayes, Kellie Hill, Diana Ho, Wendy Iwaszuk,
  Smriti Jha, Dan Klein, Jayant Krishnamurthy, Theo Lanman, Percy Liang,
  Christopher~H. Lin, Ilya Lintsbakh, Andy McGovern, Aleksandr Nisnevich, Adam
  Pauls, Dmitrij Petters, Brent Read, Dan Roth, Subhro Roy, Jesse Rusak, Beth
  Short, Div Slomin, Ben Snyder, Stephon Striplin, Yu~Su, Zachary Tellman, Sam
  Thomson, Andrei Vorobev, Izabela Witoszko, Jason Wolfe, Abby Wray, Yuchen
  Zhang, and Alexander Zotov. 2020.
\newblock \href {https://doi.org/10.1162/tacl_a_00333} {Task-oriented dialogue
  as dataflow synthesis}.
\newblock \emph{Transactions of the Association for Computational Linguistics},
  8:556--571.

\bibitem[{Bogin et~al.(2022)Bogin, Gupta, and Berant}]{bogin2022unobserved}
Ben Bogin, Shivanshu Gupta, and Jonathan Berant. 2022.
\newblock Unobserved local structures make compositional generalization hard.

\bibitem[{Bogin et~al.(2021{\natexlab{a}})Bogin, Gupta, Gardner, and
  Berant}]{bogin-etal-2021-covr}
Ben Bogin, Shivanshu Gupta, Matt Gardner, and Jonathan Berant.
  2021{\natexlab{a}}.
\newblock \href {https://doi.org/10.18653/v1/2021.emnlp-main.774} {{COVR}: A
  test-bed for visually grounded compositional generalization with real
  images}.
\newblock In \emph{Proceedings of the 2021 Conference on Empirical Methods in
  Natural Language Processing}, pages 9824--9846, Online and Punta Cana,
  Dominican Republic. Association for Computational Linguistics.

\bibitem[{Bogin et~al.(2021{\natexlab{b}})Bogin, Subramanian, Gardner, and
  Berant}]{bogin-etal-2021-latent}
Ben Bogin, Sanjay Subramanian, Matt Gardner, and Jonathan Berant.
  2021{\natexlab{b}}.
\newblock \href {https://doi.org/10.1162/tacl_a_00361} {Latent compositional
  representations improve systematic generalization in grounded question
  answering}.
\newblock \emph{Transactions of the Association for Computational Linguistics},
  9:195--210.

\bibitem[{Bras et~al.(2020)Bras, Swayamdipta, Bhagavatula, Zellers, Peters,
  Sabharwal, and Choi}]{bras-etal-2020-adversarial}
Ronan~Le Bras, Swabha Swayamdipta, Chandra Bhagavatula, Rowan Zellers,
  Matthew~E. Peters, Ashish Sabharwal, and Yejin Choi. 2020.
\newblock \href {http://proceedings.mlr.press/v119/bras20a.html} {Adversarial
  filters of dataset biases}.
\newblock In \emph{Proceedings of the 37th International Conference on Machine
  Learning, {ICML} 2020, 13-18 July 2020, Virtual Event}, volume 119 of
  \emph{Proceedings of Machine Learning Research}, pages 1078--1088. {PMLR}.

\bibitem[{Brown et~al.(2020)Brown, Mann, Ryder, Subbiah, Kaplan, Dhariwal,
  Neelakantan, Shyam, Sastry, Askell, Agarwal, Herbert{-}Voss, Krueger,
  Henighan, Child, Ramesh, Ziegler, Wu, Winter, Hesse, Chen, Sigler, Litwin,
  Gray, Chess, Clark, Berner, McCandlish, Radford, Sutskever, and
  Amodei}]{brown-etal-2020-fewshot}
Tom~B. Brown, Benjamin Mann, Nick Ryder, Melanie Subbiah, Jared Kaplan,
  Prafulla Dhariwal, Arvind Neelakantan, Pranav Shyam, Girish Sastry, Amanda
  Askell, Sandhini Agarwal, Ariel Herbert{-}Voss, Gretchen Krueger, Tom
  Henighan, Rewon Child, Aditya Ramesh, Daniel~M. Ziegler, Jeffrey Wu, Clemens
  Winter, Christopher Hesse, Mark Chen, Eric Sigler, Mateusz Litwin, Scott
  Gray, Benjamin Chess, Jack Clark, Christopher Berner, Sam McCandlish, Alec
  Radford, Ilya Sutskever, and Dario Amodei. 2020.
\newblock \href
  {https://proceedings.neurips.cc/paper/2020/hash/1457c0d6bfcb4967418bfb8ac142f64a-Abstract.html}
  {Language models are few-shot learners}.
\newblock In \emph{Advances in Neural Information Processing Systems 33: Annual
  Conference on Neural Information Processing Systems 2020, NeurIPS 2020,
  December 6-12, 2020, virtual}.

\bibitem[{Campagna et~al.(2020)Campagna, Foryciarz, Moradshahi, and
  Lam}]{campagna-etal-2020-zero}
Giovanni Campagna, Agata Foryciarz, Mehrad Moradshahi, and Monica Lam. 2020.
\newblock \href {https://doi.org/10.18653/v1/2020.acl-main.12} {Zero-shot
  transfer learning with synthesized data for multi-domain dialogue state
  tracking}.
\newblock In \emph{Proceedings of the 58th Annual Meeting of the Association
  for Computational Linguistics}, pages 122--132, Online. Association for
  Computational Linguistics.

\bibitem[{Campagna et~al.(2019)Campagna, Xu, Moradshahi, Socher, and
  Lam}]{campagna2019genie}
Giovanni Campagna, Silei Xu, Mehrad Moradshahi, Richard Socher, and Monica~S.
  Lam. 2019.
\newblock \href {https://doi.org/10.1145/3314221.3314594} {Genie: A generator
  of natural language semantic parsers for virtual assistant commands}.
\newblock In \emph{Proceedings of the 40th ACM SIGPLAN Conference on
  Programming Language Design and Implementation}, PLDI 2019, page 394–410,
  New York, NY, USA. Association for Computing Machinery.

\bibitem[{Chen et~al.(2020)Chen, Liang, Yu, Song, and
  Zhou}]{chen2020compositional}
Xinyun Chen, Chen Liang, Adams~Wei Yu, Dawn Song, and Denny Zhou. 2020.
\newblock \href
  {https://proceedings.neurips.cc/paper/2020/hash/12b1e42dc0746f22cf361267de07073f-Abstract.html}
  {Compositional generalization via neural-symbolic stack machines}.
\newblock In \emph{Advances in Neural Information Processing Systems 33: Annual
  Conference on Neural Information Processing Systems 2020, NeurIPS 2020,
  December 6-12, 2020, virtual}.

\bibitem[{Conklin et~al.(2021)Conklin, Wang, Smith, and
  Titov}]{conklin-etal-2021-meta}
Henry Conklin, Bailin Wang, Kenny Smith, and Ivan Titov. 2021.
\newblock \href {https://doi.org/10.18653/v1/2021.acl-long.258} {Meta-learning
  to compositionally generalize}.
\newblock In \emph{Proceedings of the 59th Annual Meeting of the Association
  for Computational Linguistics and the 11th International Joint Conference on
  Natural Language Processing (Volume 1: Long Papers)}, pages 3322--3335,
  Online. Association for Computational Linguistics.

\bibitem[{Csord{\'a}s et~al.(2021)Csord{\'a}s, Irie, and
  Schmidhuber}]{csordas-etal-2021-devil}
R{\'o}bert Csord{\'a}s, Kazuki Irie, and Juergen Schmidhuber. 2021.
\newblock \href {https://doi.org/10.18653/v1/2021.emnlp-main.49} {The devil is
  in the detail: Simple tricks improve systematic generalization of
  transformers}.
\newblock In \emph{Proceedings of the 2021 Conference on Empirical Methods in
  Natural Language Processing}, pages 619--634, Online and Punta Cana,
  Dominican Republic. Association for Computational Linguistics.

\bibitem[{Dahl et~al.(1994)Dahl, Bates, Brown, Fisher, Hunicke-Smith, Pallett,
  Pao, Rudnicky, and Shriberg}]{dahl-etal-1994-expanding}
Deborah~A. Dahl, Madeleine Bates, Michael Brown, William Fisher, Kate
  Hunicke-Smith, David Pallett, Christine Pao, Alexander Rudnicky, and
  Elizabeth Shriberg. 1994.
\newblock \href {https://aclanthology.org/H94-1010} {Expanding the scope of the
  {ATIS} task: The {ATIS}-3 corpus}.
\newblock In \emph{{H}uman {L}anguage {T}echnology: Proceedings of a Workshop
  held at {P}lainsboro, {N}ew {J}ersey, {M}arch 8-11, 1994}.

\bibitem[{D'Amour et~al.(2020)D'Amour, Heller, Moldovan, Adlam, Alipanahi,
  Beutel, Chen, Deaton, Eisenstein, Hoffman, Hormozdiari, Houlsby, Hou, Jerfel,
  Karthikesalingam, Lucic, Ma, McLean, Mincu, Mitani, Montanari, Nado,
  Natarajan, Nielson, Osborne, Raman, Ramasamy, Sayres, Schrouff, Seneviratne,
  Sequeira, Suresh, Veitch, Vladymyrov, Wang, Webster, Yadlowsky, Yun, Zhai,
  and Sculley}]{damour-etal-2020-underspecification}
Alexander D'Amour, Katherine Heller, Dan Moldovan, Ben Adlam, Babak Alipanahi,
  Alex Beutel, Christina Chen, Jonathan Deaton, Jacob Eisenstein, Matthew~D.
  Hoffman, Farhad Hormozdiari, Neil Houlsby, Shaobo Hou, Ghassen Jerfel, Alan
  Karthikesalingam, Mario Lucic, Yian Ma, Cory McLean, Diana Mincu, Akinori
  Mitani, Andrea Montanari, Zachary Nado, Vivek Natarajan, Christopher Nielson,
  Thomas~F. Osborne, Rajiv Raman, Kim Ramasamy, Rory Sayres, Jessica Schrouff,
  Martin Seneviratne, Shannon Sequeira, Harini Suresh, Victor Veitch, Max
  Vladymyrov, Xuezhi Wang, Kellie Webster, Steve Yadlowsky, Taedong Yun,
  Xiaohua Zhai, and D.~Sculley. 2020.
\newblock \href {https://arxiv.org/abs/2011.03395} {Underspecification presents
  challenges for credibility in modern machine learning}.

\bibitem[{Finegan-Dollak et~al.(2018)Finegan-Dollak, Kummerfeld, Zhang,
  Ramanathan, Sadasivam, Zhang, and Radev}]{finegan-dollak-etal-2018-improving}
Catherine Finegan-Dollak, Jonathan~K. Kummerfeld, Li~Zhang, Karthik Ramanathan,
  Sesh Sadasivam, Rui Zhang, and Dragomir Radev. 2018.
\newblock \href {https://doi.org/10.18653/v1/P18-1033} {Improving text-to-{SQL}
  evaluation methodology}.
\newblock In \emph{Proceedings of the 56th Annual Meeting of the Association
  for Computational Linguistics (Volume 1: Long Papers)}, pages 351--360,
  Melbourne, Australia. Association for Computational Linguistics.

\bibitem[{Fodor and Pylyshyn(1988)}]{fodor1988connectionism}
Jerry~A Fodor and Zenon~W Pylyshyn. 1988.
\newblock Connectionism and cognitive architecture: A critical analysis.
\newblock \emph{Cognition}, 28(1-2):3--71.

\bibitem[{Gal et~al.(2017)Gal, Islam, and
  Ghahramani}]{yarin-bayesian-active-learning}
Yarin Gal, Riashat Islam, and Zoubin Ghahramani. 2017.
\newblock \href {http://proceedings.mlr.press/v70/gal17a.html} {Deep bayesian
  active learning with image data}.
\newblock In \emph{Proceedings of the 34th International Conference on Machine
  Learning, {ICML} 2017, Sydney, NSW, Australia, 6-11 August 2017}, volume~70
  of \emph{Proceedings of Machine Learning Research}, pages 1183--1192. {PMLR}.

\bibitem[{Geirhos et~al.(2020)Geirhos, Jacobsen, Michaelis, Zemel, Brendel,
  Bethge, and Wichmann}]{geirhos-etal-2020-shortcut-learning}
Robert Geirhos, Jörn-Henrik Jacobsen, Claudio Michaelis, Richard Zemel,
  Wieland Brendel, Matthias Bethge, and Felix~A. Wichmann. 2020.
\newblock \href {https://doi.org/10.1038/s42256-020-00257-z} {Shortcut learning
  in deep neural networks}.
\newblock \emph{Nature Machine Intelligence}, 2(11):665--673.

\bibitem[{Gordon et~al.(2020)Gordon, Lopez{-}Paz, Baroni, and
  Bouchacourt}]{Gordon2020Permutation}
Jonathan Gordon, David Lopez{-}Paz, Marco Baroni, and Diane Bouchacourt. 2020.
\newblock \href {https://openreview.net/forum?id=SylVNerFvr} {Permutation
  equivariant models for compositional generalization in language}.
\newblock In \emph{8th International Conference on Learning Representations,
  {ICLR} 2020, Addis Ababa, Ethiopia, April 26-30, 2020}. OpenReview.net.

\bibitem[{Guo et~al.(2020)Guo, Kim, and Rush}]{guo-etal-2020-sequence}
Demi Guo, Yoon Kim, and Alexander Rush. 2020.
\newblock \href {https://doi.org/10.18653/v1/2020.emnlp-main.447}
  {Sequence-level mixed sample data augmentation}.
\newblock In \emph{Proceedings of the 2020 Conference on Empirical Methods in
  Natural Language Processing (EMNLP)}, pages 5547--5552, Online. Association
  for Computational Linguistics.

\bibitem[{Guo et~al.(2021)Guo, Zhu, Lin, Chen, Lou, and
  Zhang}]{guo2021revisiting}
Yinuo Guo, Hualei Zhu, Zeqi Lin, Bei Chen, Jian{-}Guang Lou, and Dongmei Zhang.
  2021.
\newblock \href {https://ojs.aaai.org/index.php/AAAI/article/view/16930}
  {Revisiting iterative back-translation from the perspective of compositional
  generalization}.
\newblock In \emph{Thirty-Fifth {AAAI} Conference on Artificial Intelligence,
  {AAAI} 2021, Thirty-Third Conference on Innovative Applications of Artificial
  Intelligence, {IAAI} 2021, The Eleventh Symposium on Educational Advances in
  Artificial Intelligence, {EAAI} 2021, Virtual Event, February 2-9, 2021},
  pages 7601--7609. {AAAI} Press.

\bibitem[{Hemphill et~al.(1990)Hemphill, Godfrey, and
  Doddington}]{Hemphill1990atis}
Charles~T. Hemphill, John~J. Godfrey, and George~R. Doddington. 1990.
\newblock \href {https://aclanthology.org/H90-1021} {The {ATIS} spoken language
  systems pilot corpus}.
\newblock In \emph{Speech and Natural Language: Proceedings of a Workshop Held
  at Hidden Valley, {P}ennsylvania, June 24-27,1990}.

\bibitem[{Herzig and Berant(2019)}]{herzig-berant-2019-dont}
Jonathan Herzig and Jonathan Berant. 2019.
\newblock \href {https://doi.org/10.18653/v1/D19-1394} {Don{'}t paraphrase,
  detect! rapid and effective data collection for semantic parsing}.
\newblock In \emph{Proceedings of the 2019 Conference on Empirical Methods in
  Natural Language Processing and the 9th International Joint Conference on
  Natural Language Processing (EMNLP-IJCNLP)}, pages 3810--3820, Hong Kong,
  China. Association for Computational Linguistics.

\bibitem[{Herzig and Berant(2021)}]{herzig-berant-2021-span}
Jonathan Herzig and Jonathan Berant. 2021.
\newblock \href {https://doi.org/10.18653/v1/2021.acl-long.74} {Span-based
  semantic parsing for compositional generalization}.
\newblock In \emph{Proceedings of the 59th Annual Meeting of the Association
  for Computational Linguistics and the 11th International Joint Conference on
  Natural Language Processing (Volume 1: Long Papers)}, pages 908--921, Online.
  Association for Computational Linguistics.

\bibitem[{Karamcheti et~al.(2021)Karamcheti, Krishna, Fei-Fei, and
  Manning}]{karamcheti-etal-2021-mind}
Siddharth Karamcheti, Ranjay Krishna, Li~Fei-Fei, and Christopher Manning.
  2021.
\newblock \href {https://doi.org/10.18653/v1/2021.acl-long.564} {Mind your
  outliers! investigating the negative impact of outliers on active learning
  for visual question answering}.
\newblock In \emph{Proceedings of the 59th Annual Meeting of the Association
  for Computational Linguistics and the 11th International Joint Conference on
  Natural Language Processing (Volume 1: Long Papers)}, pages 7265--7281,
  Online. Association for Computational Linguistics.

\bibitem[{Keysers et~al.(2020)Keysers, Sch{\"{a}}rli, Scales, Buisman, Furrer,
  Kashubin, Momchev, Sinopalnikov, Stafiniak, Tihon, Tsarkov, Wang, van Zee,
  and Bousquet}]{keysers2020measuring}
Daniel Keysers, Nathanael Sch{\"{a}}rli, Nathan Scales, Hylke Buisman, Daniel
  Furrer, Sergii Kashubin, Nikola Momchev, Danila Sinopalnikov, Lukasz
  Stafiniak, Tibor Tihon, Dmitry Tsarkov, Xiao Wang, Marc van Zee, and Olivier
  Bousquet. 2020.
\newblock \href {https://openreview.net/forum?id=SygcCnNKwr} {Measuring
  compositional generalization: {A} comprehensive method on realistic data}.
\newblock In \emph{8th International Conference on Learning Representations,
  {ICLR} 2020, Addis Ababa, Ethiopia, April 26-30, 2020}. OpenReview.net.

\bibitem[{Kiela et~al.(2021)Kiela, Bartolo, Nie, Kaushik, Geiger, Wu, Vidgen,
  Prasad, Singh, Ringshia, Ma, Thrush, Riedel, Waseem, Stenetorp, Jia, Bansal,
  Potts, and Williams}]{kiela-etal-2021-dynabench}
Douwe Kiela, Max Bartolo, Yixin Nie, Divyansh Kaushik, Atticus Geiger,
  Zhengxuan Wu, Bertie Vidgen, Grusha Prasad, Amanpreet Singh, Pratik Ringshia,
  Zhiyi Ma, Tristan Thrush, Sebastian Riedel, Zeerak Waseem, Pontus Stenetorp,
  Robin Jia, Mohit Bansal, Christopher Potts, and Adina Williams. 2021.
\newblock \href {https://doi.org/10.18653/v1/2021.naacl-main.324} {Dynabench:
  Rethinking benchmarking in {NLP}}.
\newblock In \emph{Proceedings of the 2021 Conference of the North American
  Chapter of the Association for Computational Linguistics: Human Language
  Technologies}, pages 4110--4124, Online. Association for Computational
  Linguistics.

\bibitem[{Kim and Linzen(2020)}]{kim-linzen-2020-cogs}
Najoung Kim and Tal Linzen. 2020.
\newblock \href {https://doi.org/10.18653/v1/2020.emnlp-main.731} {{COGS}: A
  compositional generalization challenge based on semantic interpretation}.
\newblock In \emph{Proceedings of the 2020 Conference on Empirical Methods in
  Natural Language Processing (EMNLP)}, pages 9087--9105, Online. Association
  for Computational Linguistics.

\bibitem[{Lake(2019)}]{lake-2019-metaseq2seq}
Brenden~M. Lake. 2019.
\newblock \href
  {https://proceedings.neurips.cc/paper/2019/hash/f4d0e2e7fc057a58f7ca4a391f01940a-Abstract.html}
  {Compositional generalization through meta sequence-to-sequence learning}.
\newblock In \emph{Advances in Neural Information Processing Systems 32: Annual
  Conference on Neural Information Processing Systems 2019, NeurIPS 2019,
  December 8-14, 2019, Vancouver, BC, Canada}, pages 9788--9798.

\bibitem[{Lake and Baroni(2018)}]{lake2018scan}
Brenden~M. Lake and Marco Baroni. 2018.
\newblock \href {http://proceedings.mlr.press/v80/lake18a.html} {Generalization
  without systematicity: On the compositional skills of sequence-to-sequence
  recurrent networks}.
\newblock In \emph{Proceedings of the 35th International Conference on Machine
  Learning, {ICML} 2018, Stockholmsm{\"{a}}ssan, Stockholm, Sweden, July 10-15,
  2018}, volume~80 of \emph{Proceedings of Machine Learning Research}, pages
  2879--2888. {PMLR}.

\bibitem[{Lake et~al.(2017)Lake, Ullman, Tenenbaum, and
  Gershman}]{lake2017building}
Brenden~M Lake, Tomer~D Ullman, Joshua~B Tenenbaum, and Samuel~J Gershman.
  2017.
\newblock Building machines that learn and think like people.
\newblock \emph{Behavioral and brain sciences}, 40.

\bibitem[{Lewis and Catlett(1994)}]{LEWIS1994148}
David~D. Lewis and Jason Catlett. 1994.
\newblock \href
  {https://doi.org/https://doi.org/10.1016/B978-1-55860-335-6.50026-X}
  {Heterogeneous uncertainty sampling for supervised learning}.
\newblock In William~W. Cohen and Haym Hirsh, editors, \emph{Machine Learning
  Proceedings 1994}, pages 148--156. Morgan Kaufmann, San Francisco (CA).

\bibitem[{Lewis et~al.(2020)Lewis, Liu, Goyal, Ghazvininejad, Mohamed, Levy,
  Stoyanov, and Zettlemoyer}]{lewis-etal-2020-bart}
Mike Lewis, Yinhan Liu, Naman Goyal, Marjan Ghazvininejad, Abdelrahman Mohamed,
  Omer Levy, Veselin Stoyanov, and Luke Zettlemoyer. 2020.
\newblock \href {https://doi.org/10.18653/v1/2020.acl-main.703} {{BART}:
  Denoising sequence-to-sequence pre-training for natural language generation,
  translation, and comprehension}.
\newblock In \emph{Proceedings of the 58th Annual Meeting of the Association
  for Computational Linguistics}, pages 7871--7880, Online. Association for
  Computational Linguistics.

\bibitem[{Linzen(2020)}]{linzen-2020-accelerate}
Tal Linzen. 2020.
\newblock \href {https://doi.org/10.18653/v1/2020.acl-main.465} {How can we
  accelerate progress towards human-like linguistic generalization?}
\newblock In \emph{Proceedings of the 58th Annual Meeting of the Association
  for Computational Linguistics}, pages 5210--5217, Online. Association for
  Computational Linguistics.

\bibitem[{Loula et~al.(2018)Loula, Baroni, and
  Lake}]{loula-etal-2018-rearranging}
Jo{\~a}o Loula, Marco Baroni, and Brenden Lake. 2018.
\newblock \href {https://doi.org/10.18653/v1/W18-5413} {Rearranging the
  familiar: Testing compositional generalization in recurrent networks}.
\newblock In \emph{Proceedings of the 2018 {EMNLP} Workshop {B}lackbox{NLP}:
  Analyzing and Interpreting Neural Networks for {NLP}}, pages 108--114,
  Brussels, Belgium. Association for Computational Linguistics.

\bibitem[{Lowell et~al.(2019)Lowell, Lipton, and
  Wallace}]{lowell-etal-2019-practical}
David Lowell, Zachary~C. Lipton, and Byron~C. Wallace. 2019.
\newblock \href {https://doi.org/10.18653/v1/D19-1003} {Practical obstacles to
  deploying active learning}.
\newblock In \emph{Proceedings of the 2019 Conference on Empirical Methods in
  Natural Language Processing and the 9th International Joint Conference on
  Natural Language Processing (EMNLP-IJCNLP)}, pages 21--30, Hong Kong, China.
  Association for Computational Linguistics.

\bibitem[{Nie et~al.(2020)Nie, Williams, Dinan, Bansal, Weston, and
  Kiela}]{nie-etal-2020-adversarial}
Yixin Nie, Adina Williams, Emily Dinan, Mohit Bansal, Jason Weston, and Douwe
  Kiela. 2020.
\newblock \href {https://doi.org/10.18653/v1/2020.acl-main.441} {Adversarial
  {NLI}: A new benchmark for natural language understanding}.
\newblock In \emph{Proceedings of the 58th Annual Meeting of the Association
  for Computational Linguistics}, pages 4885--4901, Online. Association for
  Computational Linguistics.

\bibitem[{Oren et~al.(2021)Oren, Herzig, and Berant}]{oren-etal-2021-finding}
Inbar Oren, Jonathan Herzig, and Jonathan Berant. 2021.
\newblock \href {https://doi.org/10.18653/v1/2021.emnlp-main.843} {Finding
  needles in a haystack: Sampling structurally-diverse training sets from
  synthetic data for compositional generalization}.
\newblock In \emph{Proceedings of the 2021 Conference on Empirical Methods in
  Natural Language Processing}, pages 10793--10809, Online and Punta Cana,
  Dominican Republic. Association for Computational Linguistics.

\bibitem[{Oren et~al.(2020)Oren, Herzig, Gupta, Gardner, and
  Berant}]{oren-etal-2020-improving}
Inbar Oren, Jonathan Herzig, Nitish Gupta, Matt Gardner, and Jonathan Berant.
  2020.
\newblock \href {https://doi.org/10.18653/v1/2020.findings-emnlp.225}
  {Improving compositional generalization in semantic parsing}.
\newblock In \emph{Findings of the Association for Computational Linguistics:
  EMNLP 2020}, pages 2482--2495, Online. Association for Computational
  Linguistics.

\bibitem[{Qiu et~al.(2022)Qiu, Shaw, Pasupat, Nowak, Linzen, Sha, and
  Toutanova}]{qiu-etal-2021-csl}
Linlu Qiu, Peter Shaw, Panupong Pasupat, Pawel Nowak, Tal Linzen, Fei Sha, and
  Kristina Toutanova. 2022.
\newblock \href {https://doi.org/10.18653/v1/2022.naacl-main.323} {Improving
  compositional generalization with latent structure and data augmentation}.
\newblock In \emph{Proceedings of the 2022 Conference of the North American
  Chapter of the Association for Computational Linguistics: Human Language
  Technologies}, pages 4341--4362, Seattle, United States. Association for
  Computational Linguistics.

\bibitem[{Sakaguchi et~al.(2020)Sakaguchi, Bras, Bhagavatula, and
  Choi}]{sakaguchi-etal-2021-winogrande}
Keisuke Sakaguchi, Ronan~Le Bras, Chandra Bhagavatula, and Yejin Choi. 2020.
\newblock \href {https://aaai.org/ojs/index.php/AAAI/article/view/6399}
  {Winogrande: An adversarial winograd schema challenge at scale}.
\newblock In \emph{The Thirty-Fourth {AAAI} Conference on Artificial
  Intelligence, {AAAI} 2020, The Thirty-Second Innovative Applications of
  Artificial Intelligence Conference, {IAAI} 2020, The Tenth {AAAI} Symposium
  on Educational Advances in Artificial Intelligence, {EAAI} 2020, New York,
  NY, USA, February 7-12, 2020}, pages 8732--8740. {AAAI} Press.

\bibitem[{Sener and Savarese(2018)}]{sener-core-set}
Ozan Sener and Silvio Savarese. 2018.
\newblock \href {https://openreview.net/forum?id=H1aIuk-RW} {Active learning
  for convolutional neural networks: {A} core-set approach}.
\newblock In \emph{6th International Conference on Learning Representations,
  {ICLR} 2018, Vancouver, BC, Canada, April 30 - May 3, 2018, Conference Track
  Proceedings}. OpenReview.net.

\bibitem[{Settles and Craven(2008)}]{settles-craven-2008-analysis}
Burr Settles and Mark Craven. 2008.
\newblock \href {https://aclanthology.org/D08-1112} {An analysis of active
  learning strategies for sequence labeling tasks}.
\newblock In \emph{Proceedings of the 2008 Conference on Empirical Methods in
  Natural Language Processing}, pages 1070--1079, Honolulu, Hawaii. Association
  for Computational Linguistics.

\bibitem[{Shaw et~al.(2021)Shaw, Chang, Pasupat, and
  Toutanova}]{shaw-etal-2021-compositional}
Peter Shaw, Ming-Wei Chang, Panupong Pasupat, and Kristina Toutanova. 2021.
\newblock \href {https://doi.org/10.18653/v1/2021.acl-long.75} {Compositional
  generalization and natural language variation: Can a semantic parsing
  approach handle both?}
\newblock In \emph{Proceedings of the 59th Annual Meeting of the Association
  for Computational Linguistics and the 11th International Joint Conference on
  Natural Language Processing (Volume 1: Long Papers)}, pages 922--938, Online.
  Association for Computational Linguistics.

\bibitem[{Tamkin et~al.(2022)Tamkin, Nguyen, Deshpande, Mu, and
  Goodman}]{tamkin-2022-active-learning}
Alex Tamkin, Dat Nguyen, Salil Deshpande, Jesse Mu, and Noah Goodman. 2022.
\newblock \href {https://arxiv.org/abs/2204.08491} {Active learning helps
  pretrained models learn the intended task}.

\bibitem[{Wallace et~al.(2019)Wallace, Rodriguez, Feng, Yamada, and
  Boyd-Graber}]{wallace-etal-2019-trick}
Eric Wallace, Pedro Rodriguez, Shi Feng, Ikuya Yamada, and Jordan Boyd-Graber.
  2019.
\newblock \href {https://doi.org/10.1162/tacl_a_00279} {Trick me if you can:
  Human-in-the-loop generation of adversarial examples for question answering}.
\newblock \emph{Transactions of the Association for Computational Linguistics},
  7:387--401.

\bibitem[{Wang et~al.(2021)Wang, Yin, Lin, and
  Xiong}]{wang-etal-2021-learning-synthesize}
Bailin Wang, Wenpeng Yin, Xi~Victoria Lin, and Caiming Xiong. 2021.
\newblock \href {https://doi.org/10.18653/v1/2021.naacl-main.220} {Learning to
  synthesize data for semantic parsing}.
\newblock In \emph{Proceedings of the 2021 Conference of the North American
  Chapter of the Association for Computational Linguistics: Human Language
  Technologies}, pages 2760--2766, Online. Association for Computational
  Linguistics.

\bibitem[{Wang et~al.(2015)Wang, Berant, and Liang}]{wang-etal-2015-building}
Yushi Wang, Jonathan Berant, and Percy Liang. 2015.
\newblock \href {https://doi.org/10.3115/v1/P15-1129} {Building a semantic
  parser overnight}.
\newblock In \emph{Proceedings of the 53rd Annual Meeting of the Association
  for Computational Linguistics and the 7th International Joint Conference on
  Natural Language Processing (Volume 1: Long Papers)}, pages 1332--1342,
  Beijing, China. Association for Computational Linguistics.

\bibitem[{Xu et~al.(2020)Xu, Campagna, Li, and Lam}]{xu2020schema2qa}
Silei Xu, Giovanni Campagna, Jian Li, and Monica~S. Lam. 2020.
\newblock \href {https://doi.org/10.1145/3340531.3411974} {Schema2qa:
  High-quality and low-cost q{\&}a agents for the structured web}.
\newblock In \emph{{CIKM} '20: The 29th {ACM} International Conference on
  Information and Knowledge Management, Virtual Event, Ireland, October 19-23,
  2020}, pages 1685--1694. {ACM}.

\bibitem[{Yin et~al.(2021)Yin, Fang, Neubig, Pauls, Platanios, Su, Thomson, and
  Andreas}]{yin-etal-2021-compositional}
Pengcheng Yin, Hao Fang, Graham Neubig, Adam Pauls, Emmanouil~Antonios
  Platanios, Yu~Su, Sam Thomson, and Jacob Andreas. 2021.
\newblock \href {https://doi.org/10.18653/v1/2021.naacl-main.225}
  {Compositional generalization for neural semantic parsing via span-level
  supervised attention}.
\newblock In \emph{Proceedings of the 2021 Conference of the North American
  Chapter of the Association for Computational Linguistics: Human Language
  Technologies}, pages 2810--2823, Online. Association for Computational
  Linguistics.

\bibitem[{Yin et~al.(2022)Yin, Wieting, Sil, and
  Neubig}]{yin-etal-2022-ingredients}
Pengcheng Yin, John Wieting, Avirup Sil, and Graham Neubig. 2022.
\newblock \href {https://doi.org/10.18653/v1/2022.acl-long.103} {On the
  ingredients of an effective zero-shot semantic parser}.
\newblock In \emph{Proceedings of the 60th Annual Meeting of the Association
  for Computational Linguistics (Volume 1: Long Papers)}, pages 1455--1474,
  Dublin, Ireland. Association for Computational Linguistics.

\end{thebibliography}

\bibliographystyle{acl_natbib}

\clearpage

\appendix
\section{Training}
\label{app:training}

Models for COVR, ATIS, Schema2QA and Overnight datasets were trained for different number of epochs depending on train set size as shown in Table \ref{tab:epochs}. Models for Schema2QA were all trained for 240 epochs. Training was run with batch sizes ranging from 8 to 20, depending on the maximum number of example tokens in each dataset and a learning rate of $3e^{-5}$ with polynomial decay. Each experiment was run with a Nvidia Titan RTX GPU and took between a few minutes to a couple of hours as we varied the training set size and number of epochs. We used exact match accuracy as our metric and following \cite{bogin2022unobserved}, we do early stopping using the test set. As our goal is to estimate the train set quality and not the model, we argue this is an acceptable choice in our setting. 

\begingroup
\begin{table}
\centering
\small
\begin{tabular}{cc}
\toprule
  Budget &  \#Epochs \\
\midrule
    50 & 160 \\
    300 & 128 \\
    600 & 96 \\
    1000 & 80 \\
\bottomrule
\end{tabular}
\caption{Number of training epochs for COVR, ATIS, Schema2QA and Overnight depending on train set size.}
\label{tab:epochs}
\end{table}
\endgroup
\section{Additional Results}
\label{app:results}

\begin{figure*}
    \centering
    \includegraphics[width=\textwidth]{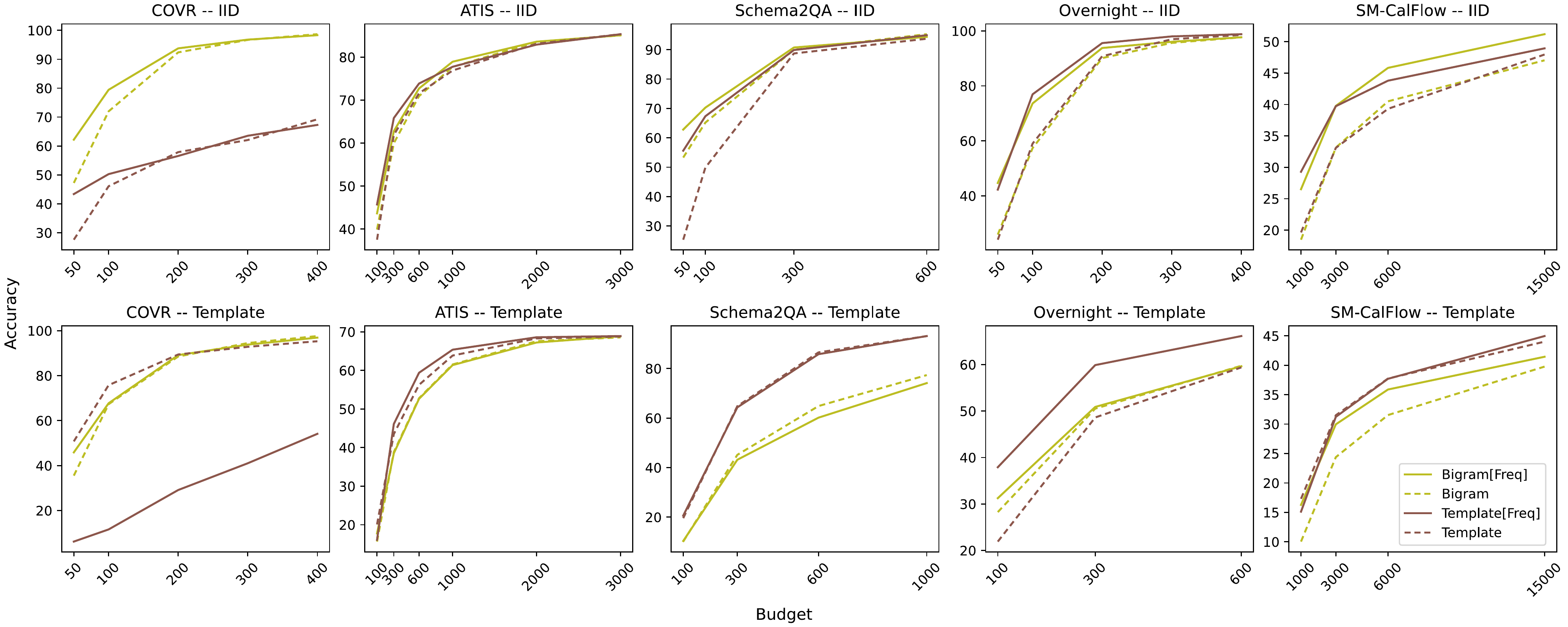}
    \caption{
    Prioritizing more frequent substructures greatly improves bigram and template diversity in IID splits with relatively minor degradation in template splits.
    }
    \label{fig:prior-efficiency}
\end{figure*}

The original Bigram and Template diversity algorithms (\bigram{} and \template{}), while good in template splits, are worse than even random subsampling in IID splits. However, replacing their substructure selection scheme with one that prioritizes frequent substructures (\bigramfreq{} and \templatefreq{}) improves both of their efficiencies in IID splits with minor degradation in template splits (Figure \ref{fig:prior-efficiency}). The only exception is the template diversity in COVR. This is expected given that it was generated from a synchronous grammar with production rules sampled uniformly at random and hence is dominated by instances with shorter templates. Thus, \templatefreq{} will only pick these. Additionally, the slightly poorer performance subtree diversity than template diversity on \smcalflow{} can be attributed to (1) imperfect program to tree conversion and (2)  the presence of free-form strings in programs which together lead to a large number of spurious subtrees.

\section{Coverage}
\label{app:coverage}

\begin{figure*}[htp]
    \centering

    \begin{subfigure}{\textwidth}
        \centering
        \includegraphics[width=\linewidth]{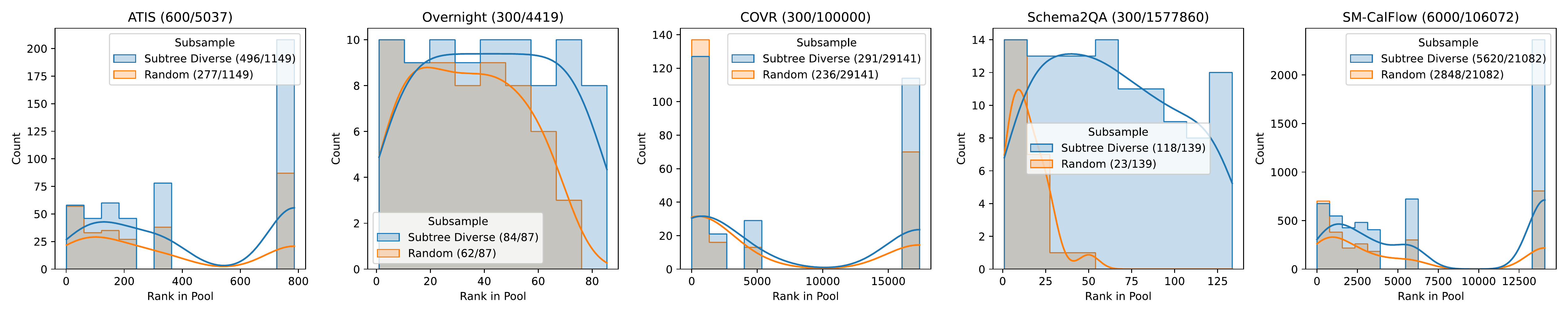}

    \end{subfigure}
    \bigskip

    \begin{subfigure}{\textwidth}
        \centering
        \includegraphics[width=\linewidth]{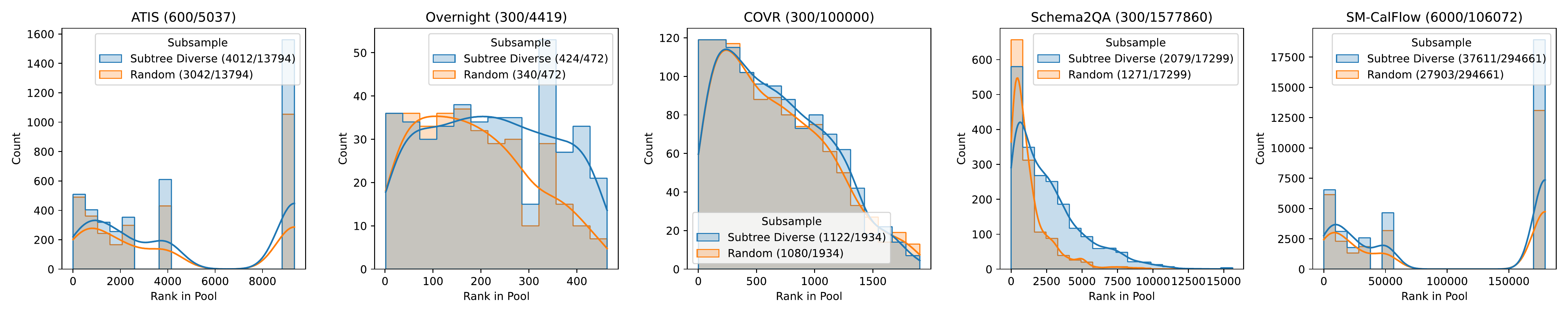}
    \end{subfigure}

    \caption{Number of unique program templates (top) and input utterance n-grams of size up to three (bottom) in diverse and random subsamples bucketed by pool ranks. Titles include size of the subsample v/s the pool while the legends include the number of unique substructures in each subsample v/s the pool.}
    \label{fig:additional-coverage}
\end{figure*}

Figure \ref{fig:additional-coverage} replicates the analysis of \S~\ref{sec:coverage} using program templates and n-grams in the input utterances as substructures showing that improving coverage over subtrees also leads to improved coverage of program templates as well as n-grams input utterances.

\end{document}